\documentclass[11pt]{article}

\usepackage[final]{acl}

\usepackage{times}
\usepackage{latexsym}
\usepackage{epigraph}

\usepackage[T1]{fontenc}

\usepackage[utf8]{inputenc}

\usepackage{microtype}

\usepackage{inconsolata}

\usepackage{graphicx}

\title{Who Warmed the Archives? LLMs Overestimate Historical Warmth}

\author{
  Claudiu Creanga\textsuperscript{2,3}\thanks{Corresponding author.},
  Liviu P. Dinu\textsuperscript{1,3} \\
  \textsuperscript{1} Faculty of Mathematics and Computer Science, \\
  \textsuperscript{2} Interdisciplinary School of Doctoral Studies, \\
  \textsuperscript{3} HLT Research Center, \\
  University of Bucharest, Romania \\
  \texttt{claudiu.creanga@s.unibuc.ro},
  \texttt{ldinu@fmi.unibuc.ro}
}

\begin{document}
\maketitle
\begin{abstract}
Historical archives are an under-used source for extending the instrumental climate record backward in time, and LLMs offer a way to extract the indices climatologists derive by hand. Beyond measuring how well systems extract this signal, we check whether their errors are safe to use for cross-century comparison, since a good correlation score does not rule out systematic, era-linked bias. Comparing lexical baselines, fine-tuned historical transformers, and LLM prompting on the Pfister temperature index across five centuries of German text, lexical methods beat every fine-tuned transformer we test, including one pretrained from scratch on historical German ($r=-0.016$). All six LLMs we test (Gemini 2.5 Flash, GPT-5-mini, DeepSeek v4 Flash, Claude Sonnet 4.6, Qwen3.7-Plus, Kimi-K2.6-Fast) show a warm bias that grows with calendar year, with the same sign in every model (slopes $+0.13$ to $+0.34$/century, $p<0.01$). The effect is modest in size ($r^2\approx0.01$--$0.05$) but consistent across six independently developed models. The best-correlated of the six, Gemini 2.5 Flash, matches the best lexical correlation ($r=0.32$) at double the error. An ablation stripping explicit dates and calendar-era markers from the quotes leaves this trend essentially unchanged, favoring an anachronistic present-day prior over the model correctly inferring the quote's era. Correlation alone is thus insufficient for vetting an LLM as a historical-climate-index oracle.
\end{abstract}

\epigraph{But where are the snows of yesteryear?}{François Villon}

\section{Introduction}

\begin{figure}[t]
\centering
\includegraphics[width=\columnwidth]{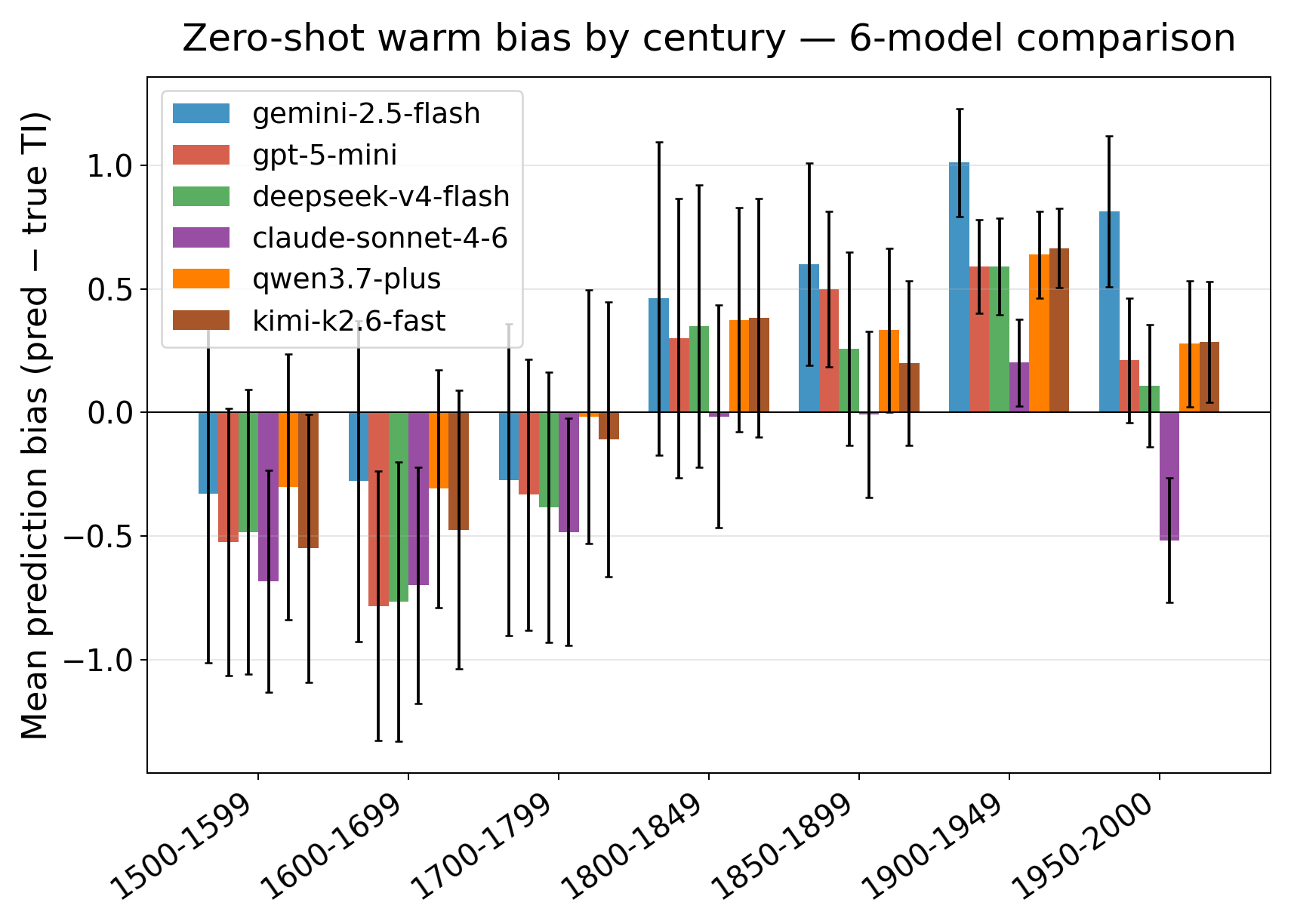}
\caption{Mean zero-shot prediction bias by century, all six audited models, identical 702-row sample: a cold bias before 1800 flips to a warm bias after, in every model.}
\label{fig:bias-by-century}
\end{figure}

Historical documentary archives (chronicles, weather diaries, and newspaper reports spanning five centuries) record climate information via the semi-quantitative ordinal indices long used in historical climatology \citep{pfister-1984-potential,glaser-riemann-2009-thousand,glaser-kahle-2020-droughts}. NLP, and increasingly LLM prompting, is an obvious way to scale up this extraction. Most evaluations of this kind stop at whether a system can extract usable climate signal from old text at all, typically measured with a correlation or F1 score. We additionally check whether that signal can be trusted for the \textbf{cross-century comparisons} a paleoclimatologist would actually want to make with it.

We compare lexical baselines, historical-domain fine-tuned transformers (Europeana BERT, GHisBERT, Historic Multilingual BERT), and zero-/few-shot LLM prompting (Gemini 2.5 Flash \citep{gemini-2025-technical}, GPT-5-mini \citep{openai-2026-gpt5}, DeepSeek v4 Flash \citep{deepseek-2026-v4}, Claude Sonnet 4.6 \citep{anthropic-2026-sonnet46}, Qwen3.7-Plus \citep{qwen-2026-plus}, Kimi-K2.6-Fast \citep{kimi-2026-k2}) on predicting the Pfister temperature index from German historical text (1500--2022), then audit whether the best LLM's errors are random or systematically structured by era across all six models. GHisBERT, pretrained on historical German from scratch, scores worse than random ($r=-0.016$) against a plain TF-IDF baseline ($r=0.32$). All six LLMs show a warm bias that grows significantly with calendar year. Gemini, the best-correlated of the six, matches that lexical correlation while carrying about twice the error (Figure~\ref{fig:bias-by-century}). Exact figures appear in \S\ref{sec:results}.

These results motivate three questions: where climate signal lives in historical text (\textbf{RQ1}), whether LLM prompting can match specialized fine-tuning (\textbf{RQ2}), and whether an LLM's miscalibration here reflects a systematic, temporally structured anachronistic prior, or is better explained as unstructured noise (\textbf{RQ3}). They yield five contributions:

\begin{itemize}
\item A controlled comparison of lexical, fine-tuned-transformer, and LLM-prompting approaches on five centuries of German/Latin text, showing lexical methods beat every fine-tuned transformer we test, including one purpose-built for this exact language variety.
\item Evidence that correlation alone can be a misleading fitness measure for an LLM's climate-index accuracy: it can tie a lexical baseline's $r$ while its absolute error runs twice as large, because its errors follow a systematic, directional pattern.
\item A cross-model audit showing this systematic error is temporally structured and replicates across \textbf{six independently developed model families} (Google, OpenAI, DeepSeek, Anthropic, Alibaba, and Moonshot AI), suggesting it is a general property of current LLMs, not an idiosyncrasy of a single model.
\item An \textbf{era-cue-stripping ablation} showing this bias survives, essentially unchanged, once explicit dates and calendar-era markers are removed from the prompt. This points to an anachronistic prior: the model does not appear to read the quote's stated date correctly and adjust its prediction accordingly.
\item A secondary case study on structured event extraction from multilingual medieval records, showing \textbf{Latin} quotes are not easier for the model than dialectal \textbf{German} despite Latin's greater standardization.
\end{itemize}

\section{Related Work}
\label{sec:related-work}

A growing line of work asks whether contemporary LLMs can be trusted to represent, or reason about, periods predating their training distribution's center of mass. \citet{underwood-etal-2025-anachronism} show that prompting or light fine-tuning does not convincingly reproduce period prose style, arguing that simulating a historical perspective may require pretraining on period corpora. \citet{cuscito-etal-2024-bert} and \citet{cuscito-etal-2025-shakespeare} similarly find a model pretrained on historical text (MacBERTh) beats general-purpose BERT and ChatGPT on Early Modern English fill-in-the-blank tasks, while the reverse holds on contemporary items, showing that prompting alone does not confer period competence. \citet{levchenko-2025-building} document a ``modernization trap'' and paradoxical ``over-historicization'' in multimodal-LLM OCR of 18th-century Russian books, arguing models treat ``old'' as a single undifferentiated space instead of a specific period. \citet{hutchinson-2024-mapping} frames such errors as historiographical, not merely as simple bugs, a challenge that echoes recent findings on how NLP systems struggle to reconcile divergent historical narratives \citep{creanga-etal-2026-rashomon}. We extend this evidence to a new modality (climate-index prediction) and setting (historical German text). Unlike \citeauthor{levchenko-2025-building}'s coarse old-vs.-new binary, our bias increases gradually across century-scale buckets, consistent with a graded sense of period.

A related literature studies what happens when a model's provided context and pretraining-derived (``parametric'') beliefs disagree. \citet{longpre-etal-2021-entity} formalize this as a \textit{knowledge conflict}, showing QA models over-rely on parametric knowledge even against contradicting context. \citet{neeman-etal-2023-disentqa} train models to disentangle context-grounded from parametric answers, and \citet{xu-etal-2024-knowledge-conflicts} survey the resulting taxonomy. \citet{li-etal-2025-taming} find the same attention heads can promote both signals simultaneously, and \citet{yamin-etal-2025-reconcile} show LLMs default almost exclusively to parametric knowledge under explicit counterfactual premises. We adopt this framing here: a quote's date and content make up the context. When the model over-predicts pre-industrial warmth for later (but still pre-1900) quotes, that pattern looks like a parametric prior that ``the world has warmed'' overriding the context.

A third literature studies LLMs' general-purpose sense of time. \citet{dhingra-etal-2022-time} show LMs trained on a fixed snapshot are miscalibrated on facts that change over time. \citet{zhao-etal-2024-set} find LMs often default to a year \textit{earlier} than their true training cutoff (``temporal chaos''), with alignment able to shift this anchor either direction. \citet{piryani-etal-2025-high} survey temporal QA more broadly. Our audit is a domain-specific instance of this problem, though our target differs from recall of time-sensitive facts: \textbf{we test whether a model's numeric climate judgment is anchored to the source quote's true century, or defaults toward a present-day prior}.

Our fine-tuned baselines draw on BERT-style models pretrained on historical text. \citet{schweter-etal-2022-hmbert} (dbmdz) release hmBERT, a multilingual BERT pretrained on OCR'd Europeana/British Library text for German, English, French, Swedish, and Finnish. We also use its earlier German-only sibling trained on noisier Europeana data (\texttt{dbmdz/bert-base-german-europeana-cased}).\footnote{Released without an accompanying paper. See \url{https://github.com/dbmdz/berts}.} \citet{beck-kollner-2023-ghisbert} instead pretrain GHisBERT from scratch on historical German back to 750 CE, arguing that continuing pretraining from a modern German checkpoint scales poorly to text this old. We include it as the baseline pretrained across our corpus's full historical depth.

The indices we predict come from historical climatology, a field that reconstructs pre-instrumental climate from documentary evidence such as chronicles and diaries. \citet{pfister-1984-potential} introduces the semi-quantitative ordinal-index approach that remains dominant in the field, converting qualitative chronicle/diary/record descriptions into a seven-point per-month scale calibrated against instrumental overlap. \citet{glaser-riemann-2009-thousand} apply this to produce the thousand-year German/Central European temperature reconstruction underlying our TI task, and \citet{glaser-kahle-2020-droughts} apply the same methodology to a Historical Humidity Index since 1500, underlying our HHI task. Both, with the underlying quotations, are hosted in tambora.org \citep{riemann-etal-2015-tambora}. Unlike this prior work, our contribution is testing whether NLP models can recover these expert-assigned indices directly from the quotations.

Finally, our audit is motivated by the growing use of LLMs within climate research itself. \citet{nabavi-etal-2026-climate} survey such applications, highlighting both the opportunities and the social, environmental, and epistemic risks they introduce. Our audit contributes one concrete instance of such a risk, systematic misjudgment of historical climate conditions, for a plausible near-term application (LLM-assisted digitization) of exactly the models we test.

\section{Data}
\label{sec:data}

All experiments use quotations from \textbf{historical German} (with occasional French and Latin) chronicles, newspapers, and administrative weather records, together with their associated documentary climate indices. The quotations and indices originate from the \textit{tambora.org} collaborative research environment for climate and environmental history \citep{riemann-etal-2015-tambora}, a database of dated, geo-referenced climate-relevant text passages linked to derived index values. The two index families we predict (a monthly Pfister-style temperature index and a monthly historical humidity index) are constructed using the semi-quantitative ordinal-index methodology of \citet{pfister-1984-potential} (\S\ref{sec:related-work}), as applied to Germany and Central Europe by \citet{glaser-riemann-2009-thousand} (temperature) and \citet{glaser-kahle-2020-droughts} (drought/humidity).

Our primary corpus pairs each quotation with its monthly Pfister temperature index (TI), an ordinal anomaly score $\mathrm{TI} \in \{-3, \dots, +3\}$ relative to a local historical baseline, which we treat as a continuous regression target under mean-squared-error loss. We use a chronological split, training on older text and evaluating on more recent text, to mimic the realistic deployment scenario of extrapolating index values into eras with progressively different source material. Table~\ref{tab:main-splits} gives split sizes and year ranges. Text is predominantly German, with historical spelling variation typical of the period (e.g.\ \textit{schney}/\textit{schnee}, \textit{fluth}/\textit{flut}). A small number of later (19th--20th century) newspaper excerpts are in French, and the earliest quotations occasionally contain Latin. We do not filter these out, so all reported numbers reflect this realistic multilingual mix. We additionally construct an expanded training variant that adds 45 further pre-1850 rows (+1\%) drawn from a larger raw extract that also contains pre-1500 quotations without a monthly TI label (see the medieval corpus described below). We report this 800-dataset variant alongside the original where relevant, since the additional rows do not meaningfully change downstream results (\S\ref{sec:results}).

\begin{table}[t]
  \centering
\begin{tabular}{lrl}
\hline
\textbf{Split} & \textbf{Records} & \textbf{Years} \\
    \hline
Train & 4{,}022 & 1500--1849 \\
Validation & 558 & 1850--1899 \\
Test (TI) & 502 & 1900--2022 \\
Test (HHI) & 666 & 1900--2026 \\
    \hline
  \end{tabular}
\caption{Main corpus splits, temperature index (TI) and humidity index (HHI). Train/validation are shared. Only the test period and size differ.}
\label{tab:main-splits}
\end{table}

To test this across the full historical range, we construct a second, purpose-built sample spanning 1500--2000 (Table~\ref{tab:audit-splits}). For the pre-1900 portion, we draw a fresh stratified sample of $n=20$ quotations per gold-TI sign (cold: $\mathrm{TI}<0$; neutral: $\mathrm{TI}=0$; warm: $\mathrm{TI}>0$) within each of five century-scale buckets (1500s--1850s), for 300 rows total, so that any cross-century difference in model bias cannot simply be an artifact of the gold-label distribution shifting across eras. For the 1900--2000 portion, we reuse the 402 held-out zero-shot test predictions from the main corpus instead of resampling, for cost efficiency when replicating the audit across six LLM providers (\S\ref{sec:experimental-setup}). This reused portion is \textit{not} sign-balanced. We treat this pre-/post-1900 sampling asymmetry as a limitation (\S\ref{sec:discussion}) and control for it throughout by reporting bias separately within each true-TI-sign subset, so that comparisons across centuries are always like-for-like on label composition.

\begin{table}[t]
\centering
\begin{tabular}{lrl}
\hline
\textbf{Portion} & \textbf{n} & \textbf{Years} \\
    \hline
Pre-1900 (stratified, fresh) & 300 & 1500--1899 \\
1900--2000 (reused test set) & 402 & 1900--2000 \\
    \hline
\textbf{Total} & \textbf{702} & 1500--2000 \\
    \hline
  \end{tabular}
\caption{Century-stratified bias-audit sample, used identically across all six audited LLMs.}
\label{tab:audit-splits}
\end{table}

As a secondary, generalization check on a second climate variable, we pair the same quotations with the monthly Historical Humidity Index (HHI $\in [-4, +4]$), which combines a historical drought index and a historical wet index derived by the same ordinal-index methodology \citep{glaser-kahle-2020-droughts}. The chronological split (Table~\ref{tab:main-splits}, HHI row) mirrors the main corpus but extends the test period to 2026 to match the available HHI series.

Because pre-1500 quotations in the raw \textit{tambora.org} extract lack a monthly TI label, we build a separate, seasonally resolved variant for the \textbf{medieval period} (1000--1499): each quotation is paired with a seasonal (winter/spring/summer/autumn) Pfister TI instead of a monthly one, split chronologically into 477 train (1000--1349), 166 validation (1350--1424), and 219 test (1425--1499) season-years. This subset is small, and the text is sparse (medieval Latin and Old/Middle/Early New High German, often with only 1--3 short quotations per season), so we read any result on this split as suggestive at best. Season labels are themselves imbalanced (mean TI: summer $+0.54$, autumn $-0.26$, spring $-0.39$, winter $-0.45$), which looks like a genuine seasonal-cycle signal in the gold data and not a sampling artifact.

As a \textbf{secondary case study} extending our signal-location question from index regression to structured information extraction, we also use a corpus of 485 unique medieval quotations (c.\ 800--1532; 104 Latin, 381 German or other) manually annotated with 2{,}769 discrete climate/hydrology/society events. Each event follows a five-field schema (category, node label, value label, location, year). The category, node label, and value label fields are each constrained to a controlled codebook of 6 categories, 68 node labels, and 33 value labels. We hold out 60 quotations (seed 42) as a fixed few-shot example pool and evaluate on the remaining 425.

\section{Experimental Setup}
\label{sec:experimental-setup}

For the main temperature-index and drought/humidity-index regression tasks, we compare four modeling approaches: lexical baselines, fine-tuned historical transformers, zero-shot LLM prompting, and few-shot LLM prompting. As lexical baselines, we fit a word-level TF-IDF vectorizer (unigrams and bigrams, minimum document frequency 3, maximum document frequency 0.8, sublinear term-frequency scaling) followed by ridge regression, with the regularization strength chosen by five-fold cross-validation over 100 log-spaced values between $10^{-3}$ and $10^{3}$. We also fit a character n-gram TF-IDF vectorizer (word-boundary-aware, n=3--5, minimum document frequency 3, maximum document frequency 0.95) followed by gradient-boosted trees (500 trees, max depth 6, learning rate 0.05, subsample and column-subsample 0.8), both with and without an additional normalized-year feature appended to the character representation. As fine-tuned transformer baselines, we take three BERT-style encoders pretrained on historical or historically adjacent German text: Europeana BERT, GHisBERT, and Historic Multilingual BERT (\S\ref{sec:related-work}). We attach a single-output regression head under mean-squared-error loss to each. All three are fine-tuned identically for 3 epochs at learning rate 3e-5, batch size 16, weight decay 0.01, with the best checkpoint selected by validation MAE.

As general-purpose LLM baselines, we prompt six models: Gemini 2.5 Flash, GPT-5-mini, DeepSeek v4 Flash, Claude Sonnet 4.6, Qwen3.7-Plus, and Kimi-K2.6-Fast. All six are queried zero-shot with an identical system prompt asking for a single Pfister-scale integer or decimal in $[-3,+3]$ given the (possibly archaic-spelled) German quote, truncated to 800 characters. The raw response is parsed for the first signed number and clamped to range. All models are queried at \texttt{temperature=0} where supported. For reasoning models, chain-of-thought tokens are discarded and only the parsed numeric prediction is kept.

We additionally test \textbf{few-shot prompting} for Gemini 2.5 Flash, drawing examples from the pre-1850 training split under two retrieval strategies (uniformly random, and nearest-neighbor by cosine similarity over the same character n-gram TF-IDF representation used for the lexical baselines), at 3-shot random, 5-shot random, and 3-shot similarity, all on the full 502-row test set. The zero-shot condition is evaluated on the 402-row reused subset described in \S\ref{sec:data}.

To probe this, we regress each model's bias (prediction minus true index) against the calendar year of the source quote, over the stratified 702-row sample described in \S\ref{sec:data}, both pooled and within each true-index-sign subset (since the pre- and post-1900 portions differ in sign-balance). We run this identically for all six models, so any difference in trends reflects only which model answered. To adjudicate between the model correctly inferring a quote's era from context and an anachronistic present-day prior applied regardless of context, we additionally construct an era-cue-stripped variant of the same 702-row sample: every explicit four-digit year (1000--2099) is replaced with a fixed placeholder, and explicit Latin/German calendar-era markers (\textit{Anno Domini}, \textit{Ao.}, \textit{n.\,Chr.}, \textit{im Jahre}, etc.) are removed. We re-run the identical zero-shot prompt on the redacted text for four of the six models (Gemini 2.5 Flash, GPT-5-mini, DeepSeek v4 Flash, and Claude Sonnet 4.6). We scope this ablation to unambiguous numeric/calendar cues, and do not attempt automatic proper-noun redaction via named-entity recognition. This is because we found that a German NER model (spaCy \texttt{de\_core\_news\_sm}) mistags common tokens in this archaic orthography badly enough (e.g.\ tagging \textit{Anno Domini} or the Latin month name \textit{Majo} as \texttt{PERSON}) that using it would inject errors in our ablation study.

For the secondary medieval event-extraction case study, we score predicted vs.\ gold events as multisets (\texttt{Counter} intersection) instead of one-to-one bipartite alignment, since events within a quote typically share one implied date. We compute year error once per quote as the absolute difference between median gold year and nearest predicted year. The codebook (the closed category/node-label/value-label vocabulary from \S\ref{sec:data}) is given verbatim in the system prompt but not enforced via schema, so codebook adherence measures genuine vocabulary reuse, since nothing forces compliance. We compare zero-shot against 3-shot random, with examples from a fixed, held-out pool of 60 quotations (seed 42).

Across all regression tasks we report MAE, RMSE, and Pearson correlation on validation and test splits. For the bias audit we report mean bias per century bucket and the slope and significance of bias regressed against year. For event extraction we report strict-match F1, category- and node-label-only F1, year MAE, location hit rate, and codebook adherence.

\section{Results}
\label{sec:results}

\begin{figure}[t]
\centering
\includegraphics[width=\columnwidth]{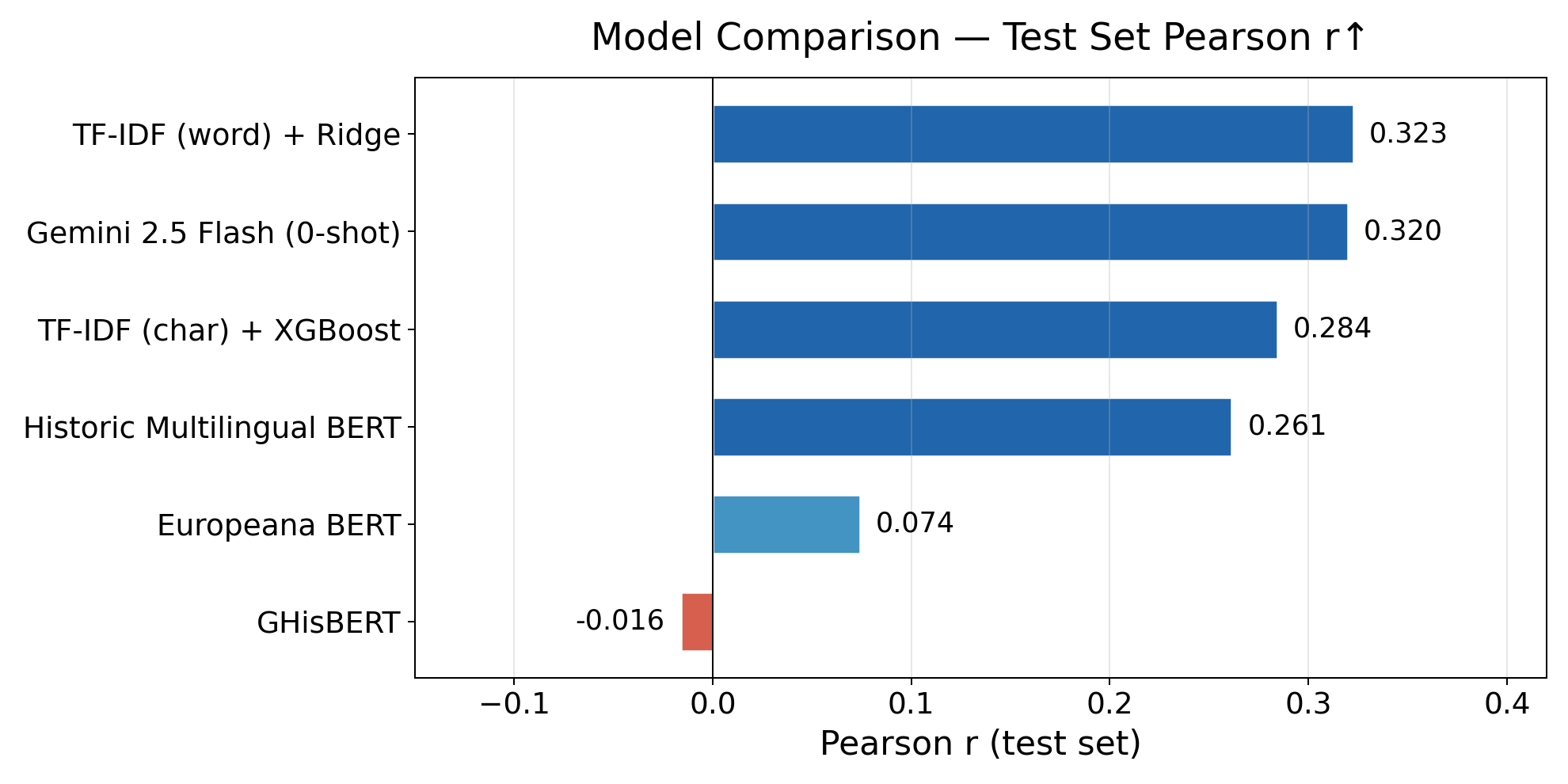}
\caption{Test-set Pearson $r$ across all main-corpus models.}
\label{fig:model-comparison}
\end{figure}

Table~\ref{tab:main-results} compares all main-corpus models on the temperature-index task. Across both the original and expanded training sets, the two TF-IDF lexical baselines outperform all three fine-tuned historical transformers on test-set correlation: the best transformer, Historic Multilingual BERT, reaches $r=0.2614$, while even the weakest lexical baseline still reaches $r=0.2771$ (char n-gram XGBoost with a year feature, expanded set), and the strongest, word-level ridge, reaches $r=0.3225$. The best absolute calibration overall belongs to the character n-gram XGBoost with a year feature on the expanded set (test MAE$=0.8596$). GHisBERT, the only baseline pretrained from scratch on historical German going back to Old High German, achieves a test correlation of $r=-0.0161$, indistinguishable from noise, despite being the domain-matched model for exactly this language variety. Figure~\ref{fig:model-comparison} visualizes this ranking directly: GHisBERT is the only model below zero, and the lexical baselines sit close to Historic Multilingual BERT well above it.

\begin{table*}[t]
\centering
\footnotesize
\resizebox{\textwidth}{!}{%
\begin{tabular}{lrrrrrr}
\hline
\textbf{Model} & \textbf{Val MAE}$\downarrow$ & \textbf{Val RMSE}$\downarrow$ & \textbf{Val r}$\uparrow$ & \textbf{Test MAE}$\downarrow$ & \textbf{Test RMSE}$\downarrow$ & \textbf{Test r}$\uparrow$ \\
\hline
TF-IDF (word) + Ridge & 0.8329 & 1.0264 & 0.3368 & 0.9012 & 1.1423 & 0.3225 \\
TF-IDF (char) + XGBoost & 0.8225 & 1.0179 & 0.3443 & 0.8687 & 1.1245 & 0.2844 \\
TF-IDF (char) + XGBoost + year & 0.8305 & 1.0260 & 0.3246 & 0.8762 & 1.1239 & 0.2905 \\
TF-IDF (word) + Ridge~[+800] & 0.8282 & 1.0219 & 0.3384 & 0.8905 & 1.1342 & 0.3091 \\
TF-IDF (char) + XGBoost~[+800] & 0.8325 & 1.0354 & 0.3096 & 0.8625 & 1.1145 & 0.2707 \\
TF-IDF (char) + XGBoost + year~[+800] & 0.8337 & 1.0348 & 0.3134 & \textbf{0.8596} & \textbf{1.1117} & 0.2771 \\
Europeana BERT & 0.8419 & -- & 0.1895 & 0.9034 & -- & 0.0741 \\
GHisBERT & 0.8467 & -- & 0.0138 & 0.8841 & -- & $-$0.0161 \\
Historic Multilingual BERT & 0.8436 & -- & 0.2740 & 0.8652 & -- & \textbf{0.2614} \\
\hline
\end{tabular}%
}
\caption{Temperature-index regression: lexical baselines (original dataset and the [+800]-expanded training variant, char n-gram = 3--5) vs.\ fine-tuned historical transformers (original dataset only).}
\label{tab:main-results}
\end{table*}

GHisBERT's test correlation remains near zero despite fine-tuning under the same regime as the other two transformers. Separately, expanding the training set by 45 rows (+1\%) via the [+800] variant produces no meaningful change in either the lexical or transformer results (e.g.\ word ridge test MAE moves from 0.9012 to 0.8905, test $r$ from 0.3225 to 0.3091).

Table~\ref{tab:fewshot} turns to whether prompting a general-purpose LLM can substitute for either the lexical baselines or the fine-tuned transformers. Gemini 2.5 Flash's zero-shot correlation on the evaluated 402-row subset ($r=0.3197$) matches the TF-IDF word ridge baseline ($r=0.3225$) almost exactly. But its MAE (1.7192) is nearly $2\times$ the ridge baseline's (0.9012), the central dissociation motivating RQ2: a metric that looks tied on correlation conceals a much larger absolute-error gap. The reason is that Gemini's errors form a right-skewed distribution of over-predictions, with mean error $+0.938$ and a tail extending to $+6$ index points on the coldest months, instead of symmetric scatter around the true value. The worst individual errors confirm the pattern: Gemini almost never predicts an index at or below $-1$ unless the quote contains unambiguous cold vocabulary (\textit{Frost}, \textit{Eis}, \textit{Schnee}). A July 1997 heatwave report and a July 1983 flood report, both true index $+3$, are predicted around $-1$, while a January 1942 ice-jam report (true index $-3$) is predicted at $+2.5$.

\begin{table}[t]
  \centering
\small
\begin{tabular}{lrrrr}
    \hline
\textbf{Condition} & \textbf{n} & \textbf{MAE}$\downarrow$ & \textbf{RMSE}$\downarrow$ & \textbf{r}$\uparrow$ \\
    \hline
Zero-shot & 402 & 1.7192 & 2.0535 & 0.3197 \\
3-shot random & 502 & 1.4975 & 1.8407 & 0.3537 \\
5-shot random & 502 & 1.4574 & 1.7897 & 0.3745 \\
3-shot similarity & 502 & 1.4404 & 1.7619 & \textbf{0.4140} \\
    \hline
  \end{tabular}
\caption{Gemini 2.5 Flash, zero-shot vs.\ few-shot, on the temperature-index test set.}
\label{tab:fewshot}
\end{table}

Few-shot prompting steadily \textbf{narrows this gap} without closing it. Moving from zero-shot to 3-shot random, 5-shot random, and 3-shot similarity-retrieved examples lowers MAE from 1.72 to 1.50, 1.46, and finally 1.44, while correlation rises from 0.32 to 0.35, 0.37, and 0.41 respectively. Similarity-based retrieval beats random retrieval at the same shot count ($r=0.4140$ vs.\ $r=0.3537$ at 3-shot), suggesting relevant grounding examples help the model recalibrate its warm bias more than example count alone. The best few-shot condition now exceeds the best fine-tuned transformer's correlation (0.4140 vs.\ 0.2614) and approaches the TF-IDF ceiling (0.3225). It still trails every lexical and transformer baseline on absolute error, though, since its MAE (1.44) remains well above even the weakest transformer's test MAE (0.90).

Table~\ref{tab:century-bias} turns the same audit in the opposite direction: instead of comparing models on a single, mostly-20th-century test period, we regress each model's zero-shot bias against the calendar year of the source quote, across the stratified 702-row sample spanning 1500--2000 (\S\ref{sec:data}), for all six models. Five of the six (Gemini, GPT-5-mini, DeepSeek v4 Flash, Qwen3.7-Plus, and Kimi-K2.6-Fast) drift from a slight cold bias on 16th--18th-century text toward an increasingly warm bias on 19th--20th-century text: Gemini moves from $-0.328$ (1500s) to $+1.011$ (1900--1949), GPT-5-mini from $-0.525$ to $+0.591$, DeepSeek from $-0.483$ to $+0.591$, Qwen3.7-Plus from $-0.300$ to $+0.638$, and Kimi-K2.6-Fast from $-0.550$ to $+0.664$, before all five ease somewhat in 1950--2000. Claude Sonnet 4.6 shows a much flatter version of the same early drift, moving from $-0.683$ in the 1500s to only $-0.017$ by 1800--1849 and never clearly turning positive in the older centuries. Unlike the other five, it then swings back to a pronounced cold bias in 1950--2000 ($-0.517$) after a brief positive bump in 1900--1949 ($+0.202$).

\begin{table}[t]
\centering
\footnotesize
\resizebox{\columnwidth}{!}{%
\begin{tabular}{lrrrrrrr}
\hline
\textbf{Century} & \textbf{n} & \textbf{Gemini} & \textbf{GPT-5} & \textbf{DeepSeek} & \textbf{Claude} & \textbf{Qwen} & \textbf{Kimi} \\
\hline
1500--1599 & 60 & $-$0.328 & $-$0.525 & $-$0.483 & $-$0.683 & $-$0.300 & $-$0.550 \\
1600--1699 & 60 & $-$0.278 & $-$0.783 & $-$0.767 & $-$0.700 & $-$0.308 & $-$0.475 \\
1700--1799 & 60 & $-$0.273 & $-$0.333 & $-$0.383 & $-$0.483 & $-$0.017 & $-$0.108 \\
1800--1849 & 60 & +0.462 & +0.300 & +0.350 & $-$0.017 & +0.375 & +0.383 \\
1850--1899 & 60 & +0.600 & +0.500 & +0.258 & $-$0.008 & +0.333 & +0.200 \\
1900--1949 & 253 & +1.011 & +0.591 & +0.591 & +0.202 & +0.638 & +0.664 \\
1950--2000 & 149 & +0.813 & +0.211 & +0.107 & $-$0.517 & +0.279 & +0.285 \\
\hline
\end{tabular}%
}
\caption{Mean zero-shot bias (predicted $-$ true index) by century, all six audited models, identical 702-row sample. ``GPT-5'' = GPT-5-mini; ``Qwen'' = Qwen3.7-Plus; ``Kimi'' = Kimi-K2.6-Fast.}
\label{tab:century-bias}
\end{table}

A linear regression of bias against year (Table~\ref{tab:bias-regression}) confirms this trend is statistically robust, not an artifact of century-bucket binning. It replicates across all six models, all on the exact same 702 quotes and system prompt ($n=702$ throughout): Gemini's bias increases by $+0.344$ index points per century ($r=0.212$, $p<0.0001$), GPT-5-mini's by $+0.272$ ($p<0.0001$), DeepSeek's by $+0.250$ ($p<0.0001$), Kimi-K2.6-Fast's by $+0.257$ ($p<0.0001$), Qwen3.7-Plus's by $+0.201$ ($p<0.0001$), and Claude's by $+0.129$ ($p=0.0054$). \textbf{Every slope is positive} and significant, pointing to a shared underlying mechanism across all six model families. Overall mean bias is not uniformly positive, though: pooled across the sample, mean bias ranges from $+0.553$ (Gemini) down to $-0.199$ for Claude, the only model net cold-biased overall while still warming significantly with year, showing the temporal drift is largely decoupled from a model's baseline calibration quality. As a case study on Gemini, splitting the regression by the true index's sign shows the effect is strongest on cold months (slope $+0.389$/century, $p=0.0007$): Gemini under-predicts cold severity progressively more for more recent centuries. It does not simply shift its mean prediction upward across the board.

\begin{table}[t]
\centering
\small
\resizebox{\columnwidth}{!}{%
\begin{tabular}{lrrr}
\hline
\textbf{Model} & \textbf{Slope/century} & \textbf{r} & \textbf{p} \\
\hline
Gemini 2.5 Flash & +0.344 & 0.212 & $<$0.0001 \\
GPT-5-mini & +0.272 & 0.199 & $<$0.0001 \\
DeepSeek v4 Flash & +0.250 & 0.177 & $<$0.0001 \\
Claude Sonnet 4.6 & +0.129 & 0.105 & 0.0054 \\
Qwen3.7-Plus & +0.201 & 0.158 & $<$0.0001 \\
Kimi-K2.6-Fast & +0.257 & 0.199 & $<$0.0001 \\
\hline
\end{tabular}%
}
\caption{Linear regression of zero-shot bias against calendar year, $n=702$ for all six models.}
\label{tab:bias-regression}
\end{table}

\begin{figure}[t]
\centering
\includegraphics[width=\columnwidth]{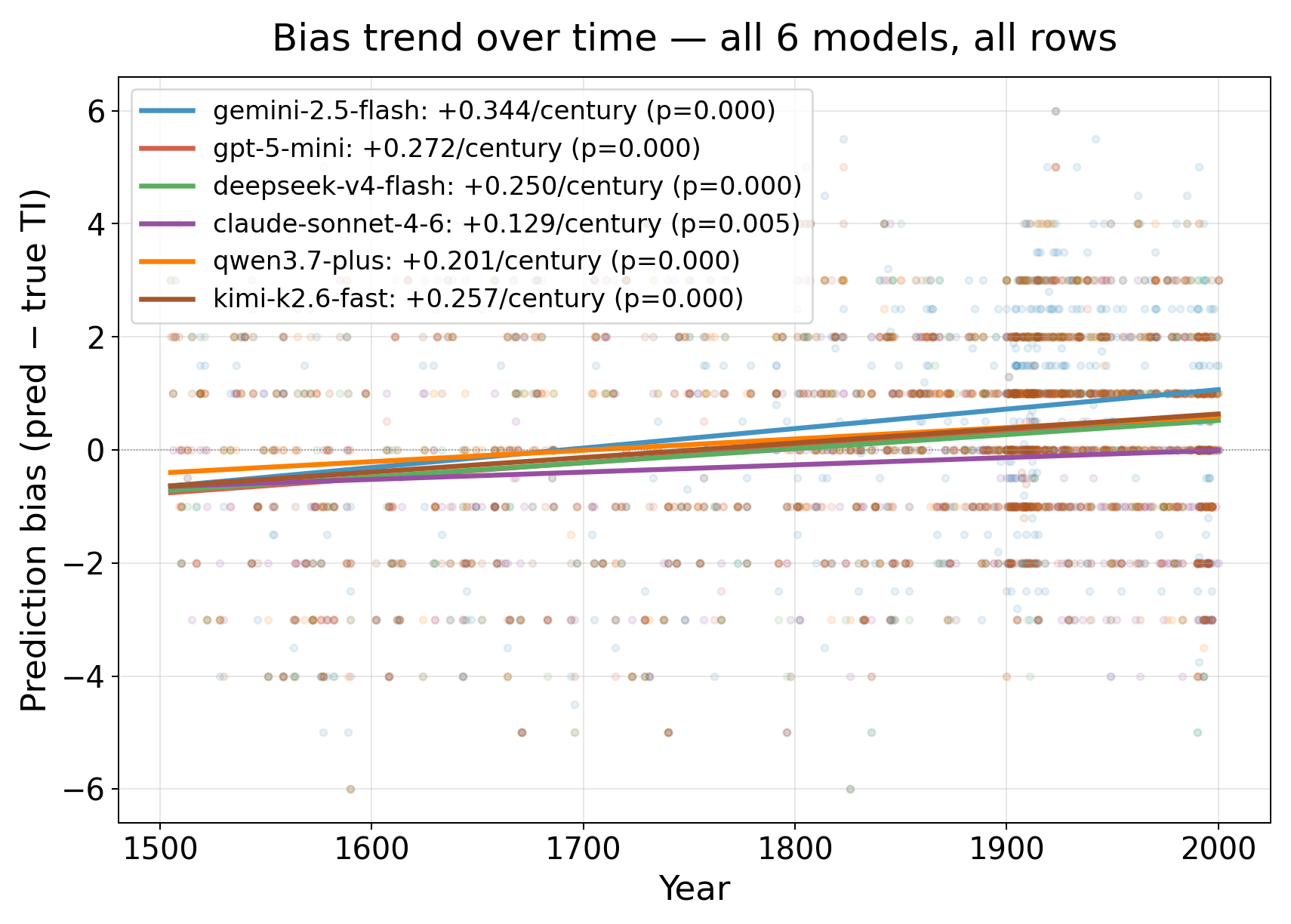}
\caption{Bias-vs-year trend, all six audited models, $n=702$ each (see Figure~\ref{fig:bias-by-century} for the corresponding per-century breakdown).}
\label{fig:bias-trend}
\end{figure}

We repeat this audit on the era-cue-stripped variant of the same 702 quotes described in \S\ref{sec:experimental-setup} (explicit years and calendar-era markers removed). Table~\ref{tab:ablation} shows the century-bias slope is essentially unchanged for every model: Gemini's slope moves from $+0.344$ to $+0.354$/century, GPT-5-mini's from $+0.272$ to $+0.269$, DeepSeek's from $+0.250$ to $+0.303$, and Claude's from $+0.129$ to $+0.124$. All remain significant ($p<0.01$, three at $p<0.0001$), with 96--122\% of the original slope surviving redaction. Since the model can no longer read an explicit date or era marker off the text, this result points to an \textbf{anachronistic present-day prior} operating largely independently of the textual date evidence that was removed. The model does not appear to be simply reading the quote's era correctly from context. We discuss two residual, unredacted era cues that could still contribute part of this surviving slope in \S\ref{sec:discussion}.

\begin{table}[t]
\centering
\small
\resizebox{\columnwidth}{!}{%
\begin{tabular}{lrrr}
\hline
\textbf{Model} & \textbf{Slope (orig.)} & \textbf{Slope (redacted)} & \textbf{p (red.)} \\
\hline
Gemini 2.5 Flash & +0.344 & +0.354 & $<$0.0001 \\
GPT-5-mini & +0.272 & +0.269 & $<$0.0001 \\
DeepSeek v4 Flash & +0.250 & +0.303 & $<$0.0001 \\
Claude Sonnet 4.6 & +0.129 & +0.124 & 0.0078 \\
\hline
\end{tabular}%
}
\caption{Bias-vs-year regression slope (points/century), original vs.\ era-redacted text, $n=702$ in both conditions.}
\label{tab:ablation}
\end{table}

Two secondary checks probe whether the lexical-beats-transformer pattern from RQ1 generalizes beyond temperature. On the drought/humidity index, the same lexical baselines reach test correlations of 0.31--0.35, comparable to temperature's 0.28--0.32 (best MAE 1.19, char XGBoost, naturally higher given HHI's wider $[-4,+4]$ scale). On the pre-1500 medieval corpus, resolved seasonally instead of monthly, word-level TF-IDF reaches a substantially higher test correlation ($r=0.4916$) despite far less training data (477 rows) and sparser, multilingual source text. This is a surprising counterpoint to the intuition that older, sparser text carries a weaker signal, though the smaller test set (219 vs.\ 502 rows) makes this only suggestive.

As a structurally different test of where historical signal lives, we extend the same zero-shot/few-shot comparison to \textbf{event extraction on the medieval corpus}. Three in-context examples roughly double Gemini's strict-match F1 from 0.1855 to 0.3834 and cut year error from 6.7 to 2.8 years, while codebook adherence stays above 99.7\% in both conditions even without schema enforcement. Contrary to the intuition that standardized Latin should be easier than fragmented medieval German dialects, node-label F1 is consistently \textit{lower} on Latin quotes, both zero-shot (0.3204 vs.\ 0.4142) and few-shot (0.4486 vs.\ 0.4989).

\section{Discussion}
\label{sec:discussion}

These results are connected. Simple lexical retrieval is hard to beat for extracting a climate signal from centuries-old text (RQ1). Prompting a general-purpose LLM with a handful of relevant examples can match or exceed even the best fine-tuned historical-domain transformer on correlation (RQ2). But zero-shot LLM errors follow a clear directional pattern: they run systematically warmer for more recent text, and this specific pattern replicates across six independently developed model families (RQ3). Together, RQ2 and RQ3 explain why a single correlation number is not enough on its own: a model can tie a lexical baseline on $r$ (RQ2) while its error is concentrated on calendar year (RQ3), the exact dimension a historian or paleoclimatologist would need to trust for cross-century comparison.

We read the \textbf{warm-bias} trend mechanistically as an instance of the parametric-versus-contextual knowledge conflict described in \S\ref{sec:related-work} \citep{longpre-etal-2021-entity,xu-etal-2024-knowledge-conflicts}. As there, the quote's date and content are the context, and the model's trained-in prior (\S\ref{sec:related-work}) overrides that context once it can roughly infer the era from vocabulary or explicit dates. This connects to the presentism/anachronism literature \citep{underwood-etal-2025-anachronism,levchenko-2025-building}. Furthermore, it highlights a broader challenge in developing models capable of adapting to diverse perspectives \citep{creanga-dinu-2024-designing}, in this case, requiring the model to adopt a temporally distinct worldview rather than defaulting to a contemporary one. Our contribution is a numeric measurement of this failure mode's cost against real, expert-derived ground truth. Correctly inferring the era and applying a reasonable prior would produce the same bias signature as drifting toward a present-day prior regardless of context, but the two carry very different implications. Our era-cue-stripping ablation (\S\ref{sec:results}) favors the latter: the slope survives, essentially unchanged, once explicit dates and calendar-era markers can no longer be read off the text.

\section{Conclusion}

We set out to test whether NLP and LLM systems can recover a physically meaningful climate signal from centuries of historical German text, and, if so, whether that signal is trustworthy enough for real cross-era scientific comparison. The answer to the first is yes: lexical methods extract this signal more reliably than every fine-tuned historical-domain transformer we test, and a general-purpose LLM can match the best lexical correlation with a handful of in-context examples. The answer to the second is more cautionary: all six LLMs we test show a systematic, era-correlated \textbf{warm bias that correlation alone does not reveal}. This bias survives an era-cue-stripping ablation run, favoring an anachronistic prior over the model correctly reading the quote's stated era. 

\textbf{Future work} should test whether this bias generalizes to other regions and languages. Also, we can test whether human annotators asked to perform the same task show a comparable presentist bias, which would suggest the model is reproducing a documented human cognitive tendency rather than exhibiting an AI-specific failure mode. Relatedly, we can check whether part of the observed drift reflects semantic change in weather vocabulary over five centuries, such as what counted as a severe frost or a mild winter shifting between 1550 and 1950, rather than the model misjudging the quote's era itself.

\section*{Limitations}

This work has several limitations. Our anachronistic-bias audit is statistically significant but explains a modest share of total prediction error ($r^2\approx0.01$--$0.05$ for the century-vs-bias trend across the six models, against an overall zero-shot MAE around 1.3--1.9 on a seven-point scale), and should not be read as the dominant source of error in LLM zero-shot climate-index prediction. The pre-1900 and post-1900 portions of our audit sample were constructed differently: one freshly stratified and sign-balanced, the other an older, unbalanced batch reused for cost efficiency. This is a sampling asymmetry we control for via sign-matched subsets but do not fully eliminate. Our era-cue-stripping ablation (\S\ref{sec:results}) weighs against the observed bias reflecting a defensible era-appropriate prior. But two residual, unredacted era cues (two-digit year abbreviations and the structured 19th/20th-century statistical-report format, \S\ref{sec:discussion}) could still contribute part of the surviving slope. We did not attempt automatic proper-noun redaction, since a quick manual check showed current historical-German NER handles this archaic orthography too poorly to trust. We test six LLM families, which we consider enough, but it doesn't confirm that the effect holds for all current LLMs. Older text in our corpus is also objectively harder to parse than modern text for reasons unrelated to anachronistic reasoning (denser archaic spelling, occasional Latin and French). This confound is hard to cleanly separate from our proposed mechanism. Finally, all data and evaluation are limited to German- (and to a lesser extent French- and Latin-) language historical sources from Central Europe. Whether the same pattern holds for other languages, regions, or climate-index conventions is untested.

\section*{Ethics Statement}

This work uses historical documentary text and derived climate indices from the \textit{tambora.org} research collaboratory \citep{riemann-etal-2015-tambora}, a publicly available academic dataset. The source material and indices describe historical weather and climate conditions and contain no personal or sensitive information about identifiable individuals. All LLM queries were made through providers' standard commercial or research APIs under their respective terms of service, and involved no human subjects. We release no new personal data. We do not anticipate direct ethical risks from this work.

\section*{Acknowledgments}

This work was partially supported by the project “Romanian Hub for Artificial Intelligence - HRIA”, Smart Growth, Digitization and Financial Instruments Program, 2021-2027, MySMIS no. 351416 and a grant of the Romanian Ministry of Research, Innovation and Digitalization, CNCS - UEFISCDI (Center of Excellence for Climate and Societal Change), project number PN-IV-P6-6.1-CoEx-2024-004, within PNCDI IV.

\bibliography{custom}

\end{document}